\documentclass{article}
\usepackage{amsmath,amsfonts,amssymb}
\usepackage{microtype}
\usepackage{graphicx}
\usepackage{booktabs} 
\DeclareMathOperator*{\argmax}{argmax}
\usepackage{wrapfig}
\usepackage{tensor}
\usepackage{xcolor}
\usepackage{soul}
\usepackage{algorithm}
\usepackage{algorithmic}
\usepackage{multicol}
\usepackage{bbold}
\usepackage{subfig}
\usepackage{amsthm}
\usepackage{amsmath}
\usepackage{mathtools}          
\usepackage{enumitem}

\newcommand{\norm}[1]{\left\lVert#1\right\rVert}

     \usepackage[preprint,nonatbib]{neurips_2020}

\usepackage[utf8]{inputenc} 
\usepackage[T1]{fontenc}    
\usepackage{hyperref}       
\usepackage{url}            
\usepackage{booktabs}       
\usepackage{amsfonts}       
\usepackage{nicefrac}       
\usepackage{microtype}      

\title{Meta-active Learning in Probabilistically-Safe Optimization}

\author{%
  Mariah L. Schrum\thanks{Use footnote for providing further information
    about author (webpage, alternative address)---\emph{not} for acknowledging
    funding agencies.} \\
  Insititute for Robots and Intelligent Machines\\
  Georgia Institute of Technology\\
  Atlanta, GA 30308 \\
  \texttt{mschrum3@gatech.edu} \\
   \And
   Mark Connolly \\
   Emory University \\
   Atlanta, GA 30308\\
   \texttt{mark.connolly@emory.edu} \\
   \AND
   Eric Cole \\
  Emory University \\
   Atlanta, GA 30308\\
   \texttt{ercole@emory.edu} \\
   \And
   Mihir Ghetiya \\
   Emory University \\
   Atlanta, GA 30308 \\
   \texttt{	mihir.v.ghetiya@emory.edu} \\
   \And
   Robert Gross \\
   Emory University \\
   Atlanta, GA 30308 \\
   \texttt{	rgross@emory.edu} \\
   \And
   Matthew Gombolay \\
   Insititute for Robots and Intelligent Machines\\
  Georgia Institute of Technology\\
  Atlanta, GA 30308 \\
  \texttt{matthew.gombolay@cc.gatech.edu} \\
}

\begin{document}

\maketitle

\begin{abstract}
Learning to control a safety-critical system with latent dynamics (e.g. for deep brain stimulation) requires taking calculated risks to gain information as efficiently as possible. To address this problem, we present a probabilistically-safe, meta-active learning approach to efficiently learn system dynamics and optimal configurations. We cast this problem as meta-learning an acquisition function, which is represented by a Long-Short Term Memory Network (LSTM) encoding sampling history. This acquisition function is meta-learned offline to learn high quality sampling strategies. We employ a mixed-integer linear program as our policy with the final, linearized layers of our LSTM acquisition function directly encoded into the objective to trade off expected information gain (e.g., improvement in the accuracy of the model of system dynamics)\color{black}~with the likelihood of safe control. We set a new state-of-the-art in active learning for control of a high-dimensional system with altered dynamics (i.e., a damaged aircraft), achieving a 46\% increase in information gain and a 20\% speedup in computation time over baselines. Furthermore, we demonstrate our system's ability to learn the optimal parameter settings for deep brain stimulation in a rat's brain while avoiding unwanted side effects (i.e., triggering seizures), outperforming prior state-of-the-art approaches with a 58\% increase in information gain. Additionally, our algorithm achieves a 97\% likelihood of terminating in a safe state while losing only 15\% of information gain.
\end{abstract}

\section{Introduction}
\label{Intro}



Safe and efficient control of a novel systems with latent dynamics is an important objective in domains from healthcare to robotics. In healthcare, deep brain stimulation devices implanted in the brain can improve memory deficits in patients with Alzheimers~\cite{posporelis2018deep} and responsive neurostimulators can counter epileptiform activity to mitigate seizures. Yet, the surgeon's manual process of finding effective inputs for each patient is arcane, time-consuming, and risky (e.g., the wrong input can trigger an ictal state or brain damage). Researchers studying \emph{active learning} and \emph{Bayesian optimization} seek to develop algorithms to efficiently and safely learn a systems' dynamics, finding some success in such healthcare applications~\cite{Ashmaig2018a,sui2018stagewise}. However, these approaches typically fail to scale up to higher-dimensional domains. As such, researchers typically only utilize simple voltage and frequency input controls for neural stimulation rather than fully modifying the waveform across 32 channels~\cite{Ashmaig2018a}. Similarly, high-dimensional tasks in robotics, such as learning the dynamics of a novel robotic systems (e.g., an autopilot learning to recover control of a damaged aircraft), require active learning methods that can succeed in these state and action spaces.\color{black}~What is critically needed then is a computational mechanism to efficiently and safely learn across low- and high-dimensional domains.

A promising area of renewed interest is that of meta-learning for online adaptation~\cite{Finn2017,Nagabandi2019,Wang2016,Andrychowicz2016}. These approaches include meta-learning of a robot controller that adapts to a new control task~\cite{Finn2017} and for fine-tuning a dynamics model during runtime ~\cite{Nagabandi2019}. However, these approaches do not consider the important problem of safely controlling a system with altered dynamics, which is a requirement for safety-critical robotic applications. Furthermore, these approaches do not actively learn. Other researchers have sought to safely and actively learn damage models via Gaussian Processes~\cite{Cully2015,Koller2019,Berkenkamp2017}, including meta-learning the Bayesian prior for a Gaussian Process~\cite{Wang2018}. However these approaches scale poorly and are sample inefficient. In this paper, we seek to overcome these key limitations of prior work by harnessing the power of meta-learning for active learning in a chance-constrained optimization framework for safe, online adaptation.

\textbf{Contributions --} We develop a probabilistically safe, meta-learning approach for active learning ("meta-active learning") that achieves state-of-the-art performance against active learning and Bayesian optimization baselines. Our acquisition function (i.e., the function that predicts the expected information gain of a data point) is meta-learned offline, allowing the policy to benefit from past experience, and provide a more robust measure of the value of an action. The key to our approach is a novel hybrid algorithm interweaving optimization via mixed-integer linear programming~\cite{schrijver1998theory} (MILP) as a policy with an acquisition function represented as a Long-Short Term Memory Network~\cite{gers1999learning} (LSTM) whose linear and piece-wise linear output layers are directly encoded into the MILP. Thus, we achieve the best of both worlds, allowing for deep learning within chance-constrained optimization. Specifically, the contributions of this paper are three-fold:\vspace{-10pt}
\begin{enumerate}[noitemsep]
    \item {Meta-active learning for autonomously synthesizing an acquisition function to efficiently infer altered or unknown system dynamics and optimize system parameters.}
    \item {Probabilistically-safe control combined with an active-learning framework through the integration of our deep learning-based acquisition function and integer linear programming.}
    \item {State-of-the art results for safe, active learning for sample-efficient reduction in model error. We achieve a 46\% increase in information gain in a high-dimensional environment of controlling a damaged aircraft, and we achieve a 58\% increase in information gain in our deep brain stimulation against our baselines.}
\end{enumerate}




\section{Preliminaries}

In this section, we review the foundations of our work in active, meta-, and reinforcement learning. 

\textbf{Active Learning} -- Labelled training data is often difficult to obtain due either to tight time constraints or lack of expert resources.  Active learning attempts to address this problem by utilizing a heuristic to quantify the amount of information an unlabelled training sample would provide the model if its label were known. A dataset $D=\{\mathcal{U_D},L_D\}$ where $\mathcal{U}$ represent a large unlabeled set of training samples, $\{x_i\}^n_{i=1}$, and $L_D$ a small set of labelled training samples, $\{x_j,y_j\}^m_{j=1}$, is available to the base learner (i.e., the algorithm used to learn the target concept), $\hat{T}_{\psi}$.  The base learner is initially trained on $L_D$ and employs a metric, $H(\mathcal{U_D})$, commonly known as an acquisition function, to determine which unlabelled sample to query for the label. The learner is then retrained on $\{L_D \cup H(\mathcal{U_D})\}$ \cite{Muslea2006,Pang2018}.

\textbf{Meta-Learning} -- Meta-learning approaches attempt to learn a method to quickly adapt to new tasks online. In contrast to active learning, meta-learning attempts to learn a skill or learning method, e.g. learning an active learning function, which can be transferable to novel tasks or scenarios. These tasks or skills are trained offline, and a common assumption is that the tasks selected at test time are drawn from the same distribution used for training
\cite{Hospedales2020}.  

\textbf{Reinforcement Learning and Q-Learning} -- A Markov decision process (MDP) is a stochastic control process for decision making and can be defined by the 5-tuple $\langle\mathcal{X},\mathcal{U},\mathcal{T},\mathcal{R},\gamma\rangle$. $\mathcal{X}$ represents the set of states and $\mathcal{U}$ the set of actions. $T: \mathcal{X} \times \mathcal{U} \times \mathcal{X}' \rightarrow [0,1]$ is the transition function that returns the probability of transitioning to state $x'$ from state $x$ applying action, $u$. $\mathcal{R}: \mathcal{X} \times \mathcal{U} \rightarrow \mathbb{R}$ is a reward function that maps a state, $x$ and action, $u$, to a reward $r \in \mathbb{R}$, and $\gamma$ weights the discounting of future rewards. Reinforcement learning seeks to synthesize a policy, $\pi: \mathcal{X} \rightarrow \mathcal{U}$, mapping states to actions in order to maximize the future expected reward. When $\pi$ is the optimal policy, $\pi^*$, the following Bellman condition holds: $Q^{\pi^*}(x,u) := \mathbb{E}_{x' \sim T}\left[R(x,u) + \gamma Q^{\pi^*}(x',\pi^*(x))\right]$ ~\cite{sutton1998}.

\subsection{Problem Set-up}
Our work is at the unique nexus of active learning, meta-learning and deep reinforcement learning with the objective of learning the Q-function as an acquisition function to describe the expected future information gained when taking action $u$ in state $x$. We define information gain as the percent decrease in the error of the objective (e.g., decrease in model error). A formal definition is provided in Supplementary\color{black}. 
The Q-function is trained via a meta-learning strategy in which an agent interacts in environments sampled from a distribution of different scenarios to distill the most effective acquisition function for active learning of system dynamics. In our context, a state $X$ is defined by the tuple $X: \langle\mathcal{U},\mathcal{L},\hat{T}_{\psi}\rangle$ where $\mathcal{U}$ consists of all possible state-action pairs and $\mathcal{L}$ is the set of state-action pairs that the agent has already experienced. $\hat{T}_{\psi}$ is a neural network function approximator of the transition dynamics, which is parameterized by $\psi$ and updated online as samples are collected. $r_i$ is proportional to the reduction of the mean squared error (MSE) loss of $\hat{T}_{\psi}$. 

\section{Safe Meta-Learning Architecture}
\begin{figure*}[t]
    \centering
    \includegraphics[width = 0.9\linewidth]{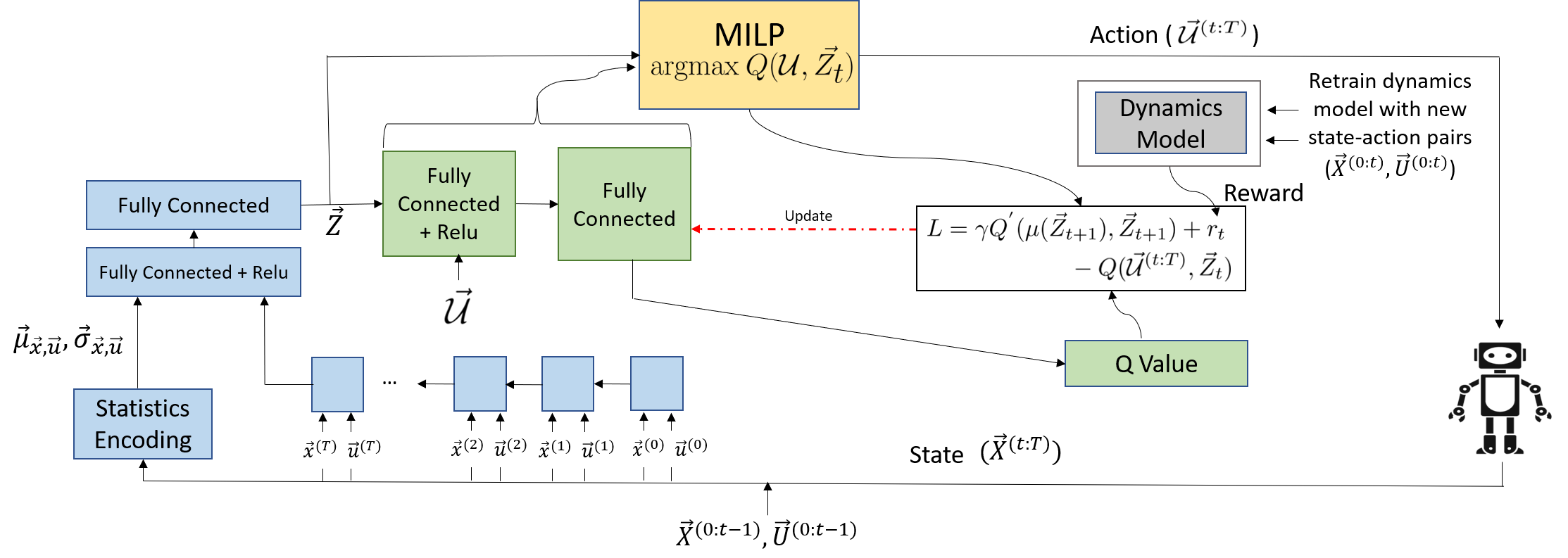}
    \caption{This figure depicts our framework.  Previously collected state-action pairs are fed into an LSTM embedding layer. Mean and variance statistics are also calculated for each state and action. The concatenated vector is fed through a fully connected layer to create the $\vec{Z}$ embedding.  Our Q-function consists of two, fully-connected, ReLU layers. Blue and green elements are the meta-learned acquisition function; the grey element is our updating dynamics model, $\hat{T}_{\psi}$, which is used to define parameters of Eq.~\ref{eq:midway3a} in the MILP; green and yellow elements are encoded as a MILP.\color{black}} 
    \label{fig:NetworkStructure}
    \vspace{-.5cm}
\end{figure*}

 Several key components are vital for learning about an unknown system in a timely manner. First, an encoding of the context of the new dynamics is important for determining where exploration should be focused and which actions may be beneficial for gaining information about the changes to the dynamics. Second, a range of prior experiences in active learning should be leveraged to best inform which actions elicit the most information in a \textbf{novel context} within a distribution of tasks. We seek to develop a framework with these key attributes to enable sample-efficient and computationally light-weight active learning. An overview of our system is shown in  Figure \ref{fig:NetworkStructure}.


\subsection{Meta-Learning Algorithm}
To infer the Q-function for an action (i.e. the acquisition function), we meta-learn over a distribution of altered dynamics as described in Algorithm \ref{alg:Overview}. For each episode, we sample from this distribution of altered dynamics and limit each episode to the number of time steps, $M$, tuned to collect enough data to accurately learn our approximate dynamics, $\hat{T}_{\psi}$, as a neural network with parameters $\psi$. We utilize Q-learning to infer the Q-function which is represented by a DQN. In our training scheme we search over the action space via a MILP, as described in Section \ref{linearProgram} and select the action set, $\vec{\mathcal{U}}^{(t:T)}$, which maximizes the Q-function while satisfying safety constraints.

The acquisition Q-function, $Q_{\theta}$, is trained via Deep Q Learning ~\cite{ganger2016double} with target network, $Q_{\phi}$, and is augmented to account for safety as described in Eq \ref{eq:obj}. The learned acquisition function, $Q_{\theta}$, is utilized by our policy (Eq. \ref{eq:obj}), which is solved via a safety-constrained MILP solver. The reward, $r_t$, for taking a set of actions in a given state is defined as the percent decrease in the MSE error of the model, $\hat{T}_{\psi}$. This Bellman loss is backpropagated through the Q-value linear and (ReLU) output layers through the LSTM encoding layers. $\mu(\vec{Z}_{t+1})$ is the set of actions, $\vec{\mathcal{U}}_{t+1}$, determined by maximizing Equation \ref{eq:obj}, which we describe in Section \ref{linearProgram}.\color{black}~The dynamics model, $\hat{T}_{\psi}$, is retrained with each new set of state-action pairs.

\begin{minipage}[t]{0.68\textwidth}
 \begin{algorithm}[H]
 \begin{algorithmic}[1]
 \small
    \caption{Meta-learning for training}\label{alg:Overview}
    \STATE {Randomly initialize $Q_{\theta}$ and $Q_{\phi}$ with weights $\theta=\phi$}
    \STATE {Initialize replay buffer, D}
    \FOR{episode=1 to N}
    \STATE Initialize $\hat{T}_{\psi}$ based on meta-learning distribution
    \STATE Collect small initial set of state-action pairs, $\mathcal{U}_0$, $\mathcal{X}_0$ \STATE Train $\hat{T}_{\psi}$ on initial set
    \FOR{i=1 to M}
    \STATE Forward pass on encoder to obtain $\vec{Z}_{i}$
    \STATE Select $\mathcal{U}_i$ from Equation \ref{eq:obj}
    \STATE Execute actions $\mathcal{U}_i + \mathcal{N}$ according to exploration noise ; observe states $\mathcal{X}_{i+1}$ 
    \STATE Retrain $\hat{T}_{\psi}$ on $\mathcal{X}_{0:T}$, $\mathcal{U}_{0:T}$ and observe reward $r_i$
    \STATE $D \leftarrow D \cup  \langle\mathcal{U}_i$,$\mathcal{X}_i$,$\mathcal{X}_{i+1},r_i\rangle$ 
    \STATE Sample a batch of transitions from D
    \STATE Perform forward pass through encoder to obtain $\vec{Z}_{t+1}$
    \STATE Calculate $y_t=r_t+\gamma Q_{\phi}\Big(\mu(\vec{Z}_{t+1}),\vec{Z}_{t+1}\Big)$ 
    \STATE Update $Q_\theta$ according to $ \Big(y_t-Q_{\theta}(\vec{\mathcal{U}}_t,\vec{Z}_{t})\Big)$
    \STATE $Q_{\phi} \leftarrow \tau Q_{\theta} + \tau(1-Q_{\phi})$
    \ENDFOR
    \ENDFOR;

\end{algorithmic}
\end{algorithm}
\end{minipage}
\hfill
\begin{minipage}[t]{0.3\textwidth}
\begin{algorithm}[H]

 \begin{algorithmic}[1]
 \small
    \caption{Meta-learning for testing}\label{alg:2}
      \STATE Draw test example from distribution of damage conditions
        \STATE Initialize $\hat{T}_{\psi}$
        \FOR{i=1 to M}
        \STATE Do forward pass through encoder to get $\vec{Z}_{i}$
        \STATE Select actions, $\mathcal{U}_i$, according to Equation \ref{eq:obj}
        \STATE Execute actions, $\mathcal{U}_i$; observe states $\mathcal{X}_{i+1}$ and reward $r_i$
        \STATE Retrain $\hat{T}_{\psi}$ on $\mathcal{X}_{i+1}$,$\mathcal{U}_{i+1}$
        \ENDFOR
        \end{algorithmic}
  \end{algorithm}
\end{minipage}

\subsection{Mixed-Integer Linear Program}
\label{linearProgram}
Our objective (Equation \ref{eq:obj}) is to maximize both the probability of the system remaining in a safe configuration and the information gained along the trajectory from $[t,t+T)$. 
\begin{align}
    \vec{\mathcal{U}}^{(t:T)^*} &=\argmax_{\vec{\mathcal{U}}^{(t:T)} \in \boldsymbol{\vec{\mathcal{U}}}^{(t:T)}}  \lambda_1 \Big[Q(\vec{\mathcal{U}}^{(t:T)},\vec{Z}) - \bar{Q}_{\theta}(\cdot,\vec{Z})\Big]+ \text{Pr}\Big\{ \norm{\vec{x}_{t+T} - \vec{x}^r}_1 \leq r \Big\} \label{eq:obj}
\end{align}\indent 
$Q(\vec{\mathcal{U}}^{(t:T)},\vec{Z})$ describes the expected information gained along the trajectory when the set of actions $\vec{\mathcal{U}}^{(t:T)}$ is taken in the context of the embedding $\vec{Z}$, and $\bar{Q}_{\theta}(\cdot,\vec{Z})$ is the expected Q-value, which we discuss further in Section \ref{eq:QFunc}. $\lambda_1$ is a hyper-parameter that can be tuned to adjust the tradeoff between safety and information gain.  We derive a linearization of this equation based on an assumption of Gaussian dynamics which can then be solved via MILP.

\begin{figure}[t]    
    \centering    
\includegraphics[width = .7\linewidth]{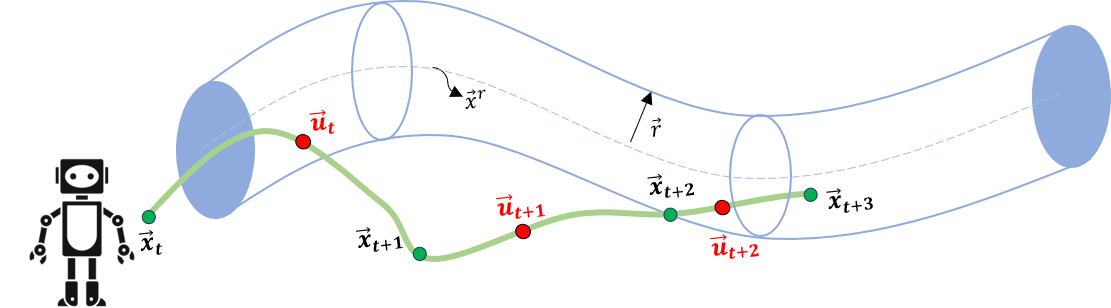} \caption{This figure depicts the d-dimensional sphere of safety. $\vec{x}^r $ denotes the safest state for the system and the shaded region is the set of all safe states.  Action, $\vec{u}_t$, is an exploratory action, which may bring the system outside of the sphere of safety. Action, $\vec{u}_{t+1}$, estimates that the system returns to a safe state with probability $1-\epsilon$. 
}     \label{fig:Cylinder}
\end{figure}

\paragraph{Definition of Safety -} We define a sphere of safety, parameterized by the safe state $x_r$ and radius $\vec{r}$ which describe all safe states of the system.  For example, in the case of an aircraft, $\vec{x}^r $ would be straight and level flight and the sphere of safety would define any deviation from level flight that is still considered safe. Additionally, we assume that our model error comes from a Gaussian distribution with known mean and variance. This assumption allows us to impose safety constraints which can either be enforced with a probability of one, or this requirement can be relaxed to increase the potential for information gain.

By assuming that our model error originates from a Gaussian distribution, we can linearize our probability constraints described in Equation \ref{eq:midway3a} to include in the MILP (See Supplementary for full derivation). Here, $\Phi^{-1}$ is the inverse cumulative distribution function for the standard normal distribution and $1-\epsilon_d$ denotes the probability level. $\sigma$ represents the uncertainty in the dynamics network parameters and $\bar{a}$ and $\bar{b}$ are the point estimates of the dynamics. $d$ represents the $d^{th}$ row and $j$ the columns. We compute the uncertainty, $\sigma$, of our network via bootstrapping \cite{Statistics2020,Franke1998}. The components of $\sigma$ are calculated as described in the Supplementary. 
\par\nobreak{\parskip0pt \small \noindent\begin{align}
    & \text{Pr}\Big\{ \norm{\vec{x}_{t+T} - \vec{x}^r}_1 \leq r \Big\} \longrightarrow \nonumber \\
    &\quad\quad\quad\quad  \Big\|\Phi^{-1}(1-\epsilon_d)\sqrt{\sum_{d^{'}}\sigma_{d,d^{'}}^2{x}_j^{(t)^2}+\sum_{j}\sigma^{^2}_{d,j}\mathcal{U}^{(t:T)^2}_{j}}+ \Gamma_k  \vec{\mathcal{U}}^{(t:T)^2} - \Delta_d^{(t:2)}\Big\|_1 <  r_d, \forall d\label{eq:midway3a}
\end{align}}Our policy aims to take a set of information rich actions at time $t$, potentially out of the d-dimensional sphere of safety, while guaranteeing with probability $1-\epsilon$ that the system will  return to a safe state in the next time step after $T$ time steps. Thus, our $T$ step propagation allows the system to deviate from the safe region, if desired, to focus on actively learning so long as it can return to the safe region with high confidence.

\subsection{An LSTM Q-Function as a Linear Program}
\label{eq:QFunc}
We leverage an LSTM as a function approximator for Q-learning. To blend our Q-function with the MILP, we pass the LSTM output embedding along with statistics of each state and action through a fully connected layer with ReLU activations.\footnote{We include the mean and standard deviation of previously collected states and actions as we find the first and second moments of the data to be helpful, additional features.} This design enables us to backprop the Bellman residual through the output layers encoded in the MILP all the way to the LSTM inputs. Thus we can leverage the power of an LSTM and mathematical programming in a seamless framework. \footnote{Details on the hyperparameter settings including learning rates and layer sizes can be found in  the Supplementary along with the derivation for the linearization of the Q-function for inclusion in the linear program.}

Given our meta-learned acquisition function and chance-constrained optimization framework, we can now perform probabilistically-safe active learning for systems such as robots to respond to damage or other augmented-dynamics scenarios. Algorithm \ref{alg:2} describes how we perform our online, safe and active learning. Intuitively, our algorithm initializes a new dynamics model to represent the unknown or altered dynamics, and we iteratively sample information rich, safe actions via our MILP policy, update our dynamics model, and repeat.

 \newtheorem{thm}{Theorem}

\section{Experimental Evaluation}
We design two experiments to validate our approach's ability to safely and actively learn altered or unknown dynamics in real time.  We compare our approach to several baseline approaches and demonstrate that our approach is superior in terms of both information gain and computation time. More details on the experimental domains can be found in Supplementary.

\subsection{Asynchronous Distributed Microelectrode Theta} 
\label{RNS}




Asynchronous distributed microelectrode theta stimulation (ADMETS) is a cutting edge deep brain stimulation approach to treating seizure conditions that cannot be controlled via pharmacological methods.  In ADMETS, a neuro-stimulator is implanted in the brain to deliver continuous electrical pulses to reduce seizures. However, there is no clear mapping from parameter values to reduction in seizures that applies to all patients, as the optimal parameter settings can depend on the placement of the device, the anatomy of an individual's brain and other confounding factors. Further, a latent subset of parameters can cause negative side-effects.

\begin{figure}[t]%
    \centering
        \subfloat[ADMETS Implantation.]{{\includegraphics[height = 2.5cm, width = .25\linewidth]{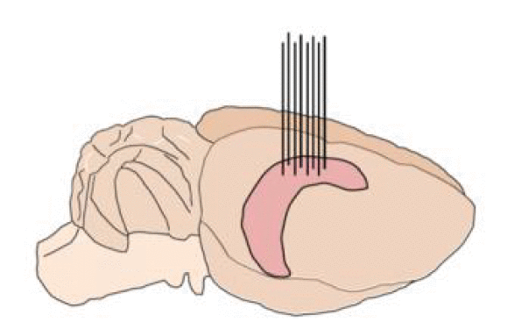}\label{fig:rat} }}%
    \qquad
        \subfloat[High-dimensional Domain; Courtesy: Flightgear.]{{\includegraphics[height = 2.5cm]{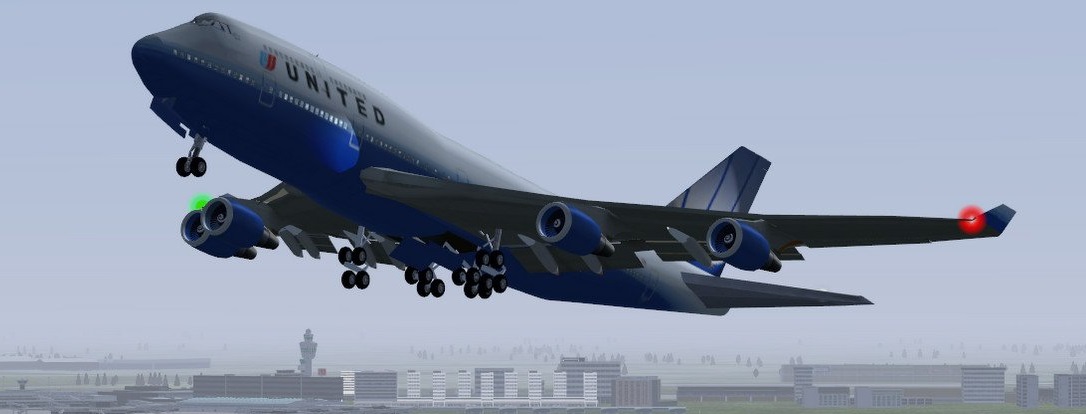}\label{fig:simulation} }}%
    \caption{Figure \ref{fig:rat} depicts a surgical implantation in a rat brain of ADMETS device \cite{Ashmaig2018a} (Section \ref{RNS}). Figure \ref{fig:simulation} depicts the high-dimensional domain for dynamical control (Section \ref{sec:Aircraft}).}
    \label{fig:example}%
\end{figure}

We test our algorithm's ability to safely determine the optimal parameter settings on a dataset consisting of a  score quantifying memory function resulting from a sweep over the parameter settings in six rat models, as shown in Fig \ref{fig:rat}. Based on the work in \cite{Ashmaig2018a}, we fit Guassian Processes to each set of data to simulate the brain of each rat. The objective is to determine the amplitude of the signal which maximizes the discrimination area (DA) without causing unwanted side effects in the rat that typically occur when the discrimination score drops below zero. Because we do not want the score to drop below zero, we define the time horizon over which the rat can leave the sphere of safety as $T=0$. The reward signal utilized by our meta-learner is the percent decrease in the optimal parameter error at each time step. The optimal parameters are defined as the parameters that maximize the discrimination area. 

\subsection{High-dimensional Domain (Recovering Dynamics of a Damaged Aircraft)}
\label{sec:Aircraft}
Active learning algorithms can be ineffective in high-dimensional domains. As such, we seek to stress-test our algorithms in just such a domain: Learning online of a damaged aircraft with nonlinear dynamics and the tight time constraints required to learn a high-quality model of the system before entering an unrecoverable configuration (e.g., spin) or crashing. We base our simulation on theoretical damage models from prior work describing the full equations of motion~\cite{Watkiss1994,Zhang2017,Ouellette2010} within the Flightgear virtual environment. The objective of this domain is to learn the difference between the altered dynamics that result from the damage and to maintain safe flight. We consider safe flight to be our designated safe state, $\vec{x}^r$. The aircraft takes an information rich action potentially resulting in a deviation outside of the d-dimensional sphere of safety.
The next action is constrained to guarantee that the plane returns to the sphere of safety with probability $1-\epsilon$ via action $u_{t+1}$. 




\subsection{Baseline Comparisons}
We include the following baselines in active learning and Bayesian optimization for our evaluation.\vspace{-5pt}
\begin{itemize}[leftmargin=*,noitemsep]
\item \textbf{Epistemic Uncertainty \cite{Hastie2017}} - This active learning metric  quantifies the uncertainty in the output of the model for each training example and chooses the action that maximizes uncertainty.

\item \textbf{Maximizing Diversity \cite{Schrum2020}} - This acquisition function selects actions which maximize the difference between previously seen states and actions.

\item \textbf{Bayesian Optimization (BaO) \cite{Ashmaig2018a}} - This algorithm was developed for the ADMETS domain (Section \ref{RNS}) and is based upon a Gaussian Process model.

\item \textbf{Meta Bayesian Optimization (Meta BO) \cite{Wang2018b}} - This approach meta-learns a Gaussian process prior offline over previously sampled data.

\item \textbf{Learning Active Learning (LAL) \cite{Konyushkova2017}} - This approach meta-learns an aquisition function, which predicts the decrease in error of the base learner offline, with hand-engineered features.
\end{itemize}\vspace{-5pt}
For our ADMETS memory optimization domain, we benchmark against \cite{Ashmaig2018a,Konyushkova2017,Wang2018b} as those baselines are designed or easily adapted to optimization tasks. For our high-dimensional domain, we likewise benchmark against \cite{Ashmaig2018a,Hastie2017,Konyushkova2017,Schrum2020} as those baselines are designed or easily adapted to system identification (i.e., model learning) tasks. We note that we do not compare to MAML\cite{Finn2017} or the approach by Nagabandi et al.~\cite{Nagabandi2019}, as these algorithms have no notion of safety or active learning.

\section{Results}
\label{Results}
We empirically validate that our meta-active learning approach outperforms baselines across both the ADMETS and high-dimensional domains in terms of its ability to actively learn latent parameters, its computation time, and its ability to safely perform these tasks.

\begin{figure}[h]%
    \centering
    \subfloat[]{{\includegraphics[height = 3cm, width = 0.32\textwidth]{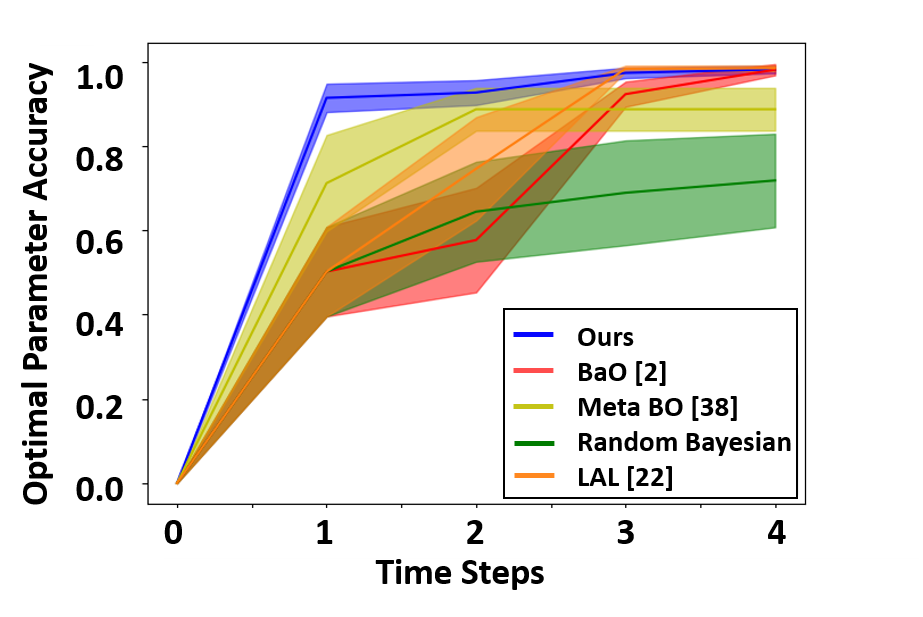}\label{fig:RNS} }}%
    \subfloat[]{{\includegraphics[height = 3cm, width = 0.32\textwidth]{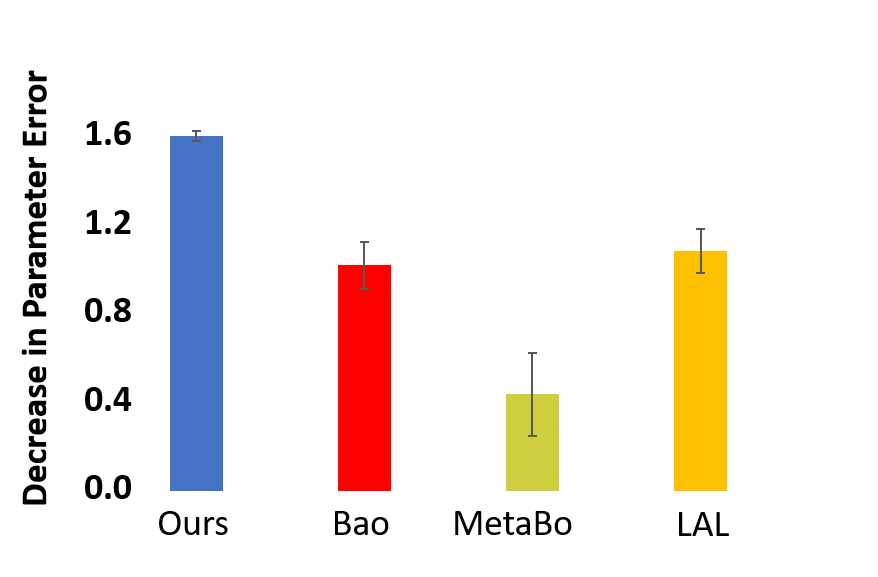}\label{fig:bargraphRNS} }}%
 \subfloat[]{{\includegraphics[height = 3cm, width = 0.32\textwidth]{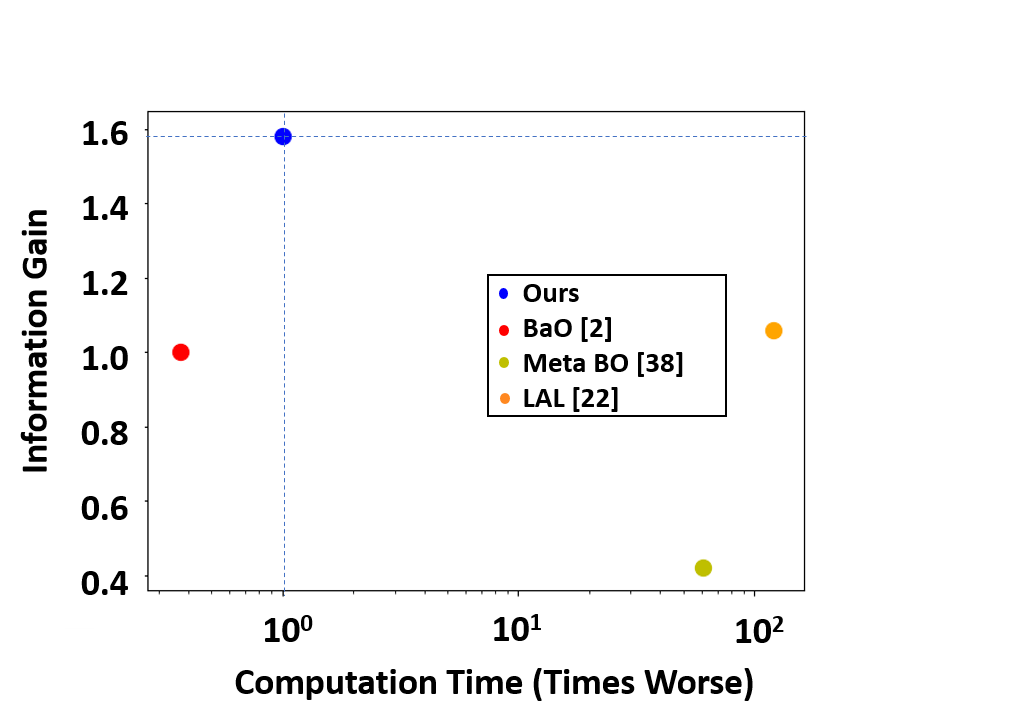}\label{fig:scatterRNS} }}%
\caption{This figure depicts the results of our empirical validation in the ADMETS domain benchmarking algorithm accuracy per time step (Fig.~\ref{fig:RNS}) and overall (Fig.~\ref{fig:bargraphRNS}) as well as accuracy versus computation time (Fig.~\ref{fig:scatterRNS}).}
\label{fig:Results1}%
\end{figure}

 \begin{figure}[b]%
    \centering
    \subfloat[]{{\includegraphics[height = 3cm, width = 0.32\textwidth]{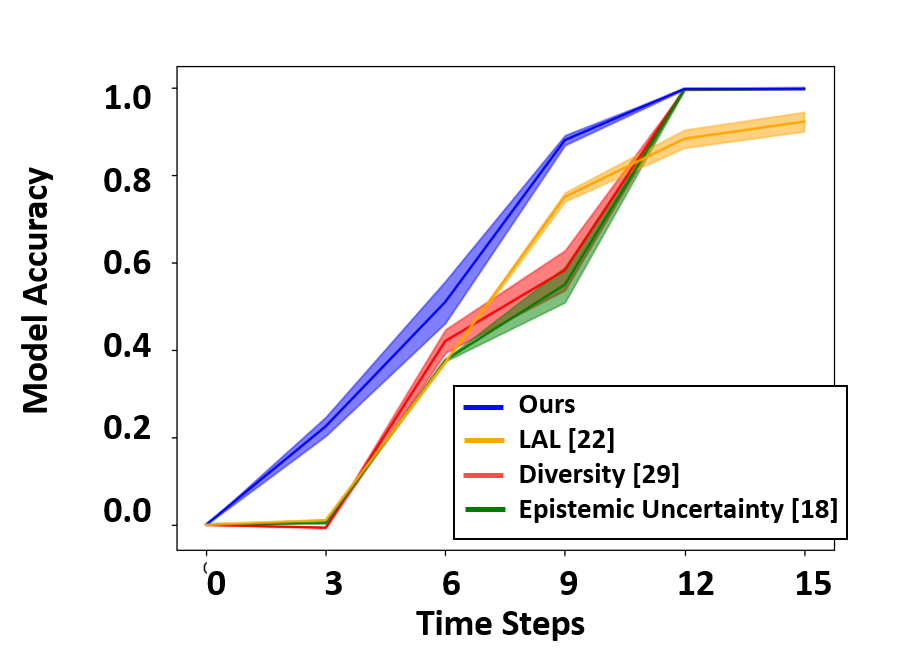}\label{fig:ALFunctions} }}%
    \subfloat[]{{\includegraphics[height = 3cm, width = 0.32\textwidth]{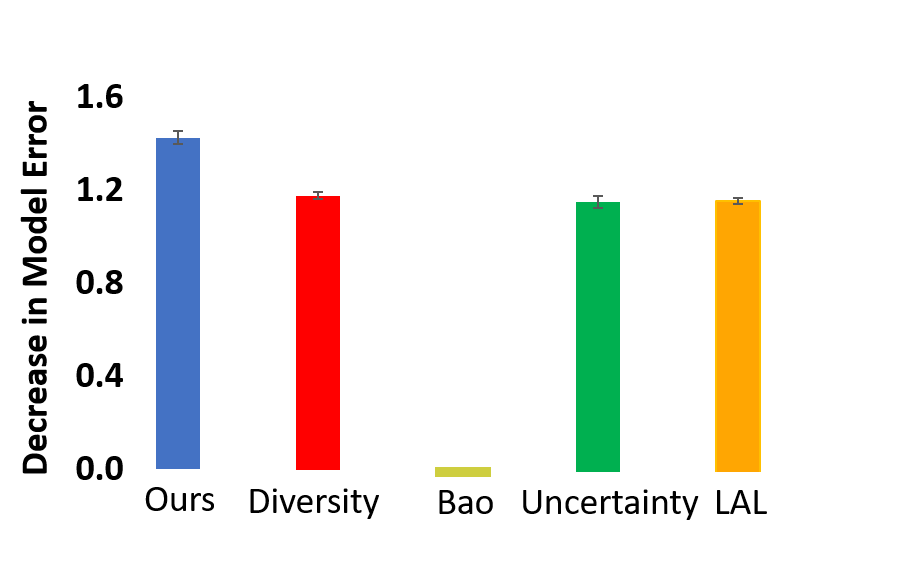}\label{fig:bargraph} }}%
    \subfloat[]{{\includegraphics[height = 3cm, width = 0.32\textwidth]{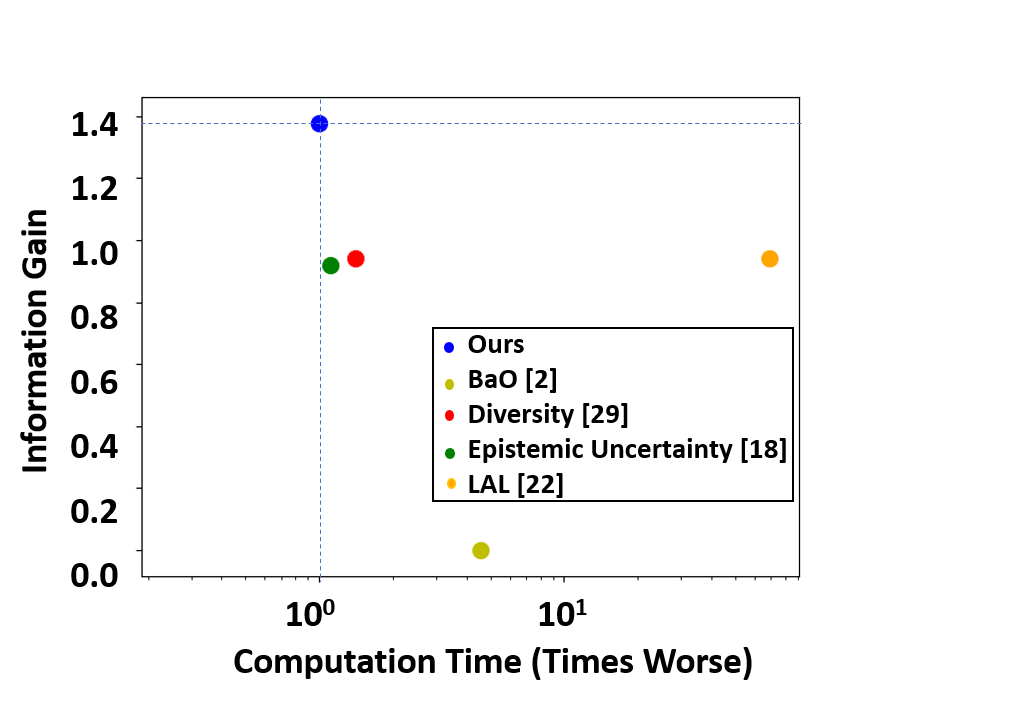}\label{fig:scatterPlane} }}%
\caption{This figure depicts the results of our empirical validation in the high-dimensional domain benchmarking algorithm accuracy per time step (Fig.~\ref{fig:ALFunctions}) and overall (Fig.~\ref{fig:bargraph}) as well as algorithm accuracy versus computation time (Fig.~\ref{fig:scatterPlane}). Error is calculated in batches of three time steps, enabling the robot to deviate from the safe region temporarily to gain information. We note that Meta BO is not included in this domain as it was designed for optimization tasks, e.g. the ADMETS domain.\color{black}}
\label{fig:Results2}%
\end{figure}

\paragraph{Active Learning --} 

Results from both the ADMETS and the high-dimensional domains empirically validate that our algorithm is able to more quickly, in terms of both computation time and information, learn the optimal control parameters of a system (Fig.~\ref{fig:RNS}) and learn a system's dynamics (Fig.~\ref{fig:ALFunctions}) while remaining safe. In the ADMETS domain, our model on average selects an action which results in a 58\% higher information gain than BaO and 87\% higher information gain than Meta BO on average. In the high-dimensional domain, Fig \ref{fig:ALFunctions} depicts the advantage of our method as a function of the number of the number of actions taken to learn the system's dynamical model; we achieve a 49\% and 46\% improvement over \cite{Hastie2017} and \cite{Schrum2020}, respectively and a 46\% improvement over \cite{Konyushkova2017} (Fig.~\ref{fig:bargraph}).



\paragraph{Computation Time --} 
We demonstrate that our method also outperforms previous work in terms of computation time. Across both domains, our approach not only achieves a more efficient reduction in model error and improvement in information gain, we are also faster than all baselines in the more challenging, high-dimensional (Fig.~\ref{fig:scatterPlane}) environment. In the lower-dimensional, ADMETS environment (Fig.~\ref{fig:scatterRNS}), BaO has a slight advantage in computation time, but our algorithm trades the time for a 58\% information gain over BaO. Additionally, we are 68x faster than LAL and 61x faster than Meta BO, the two other meta-learning approaches we benchmark against.

\paragraph{Safety --}
We empirically validate that we achieve an 87\% probability that the aircraft will return to the sphere of safety while greatly increasing information gain at each time step as shown in Fig \ref{fig:ALFunctions}, compared with 99.9\% safe return to the sphere when maximizing safety (i.e., $\lambda_1 \leftarrow 0$, resulting in passive learning). Furthermore, in the ADMETS domain, we are able to guarantee with a probability of 97\% that seizures will not occur while only losing 15\% information gain compared to when no safety constraints are imposed.

\subsection{Discussion}
Through our empirical investigation, we have demonstrated that our meta-learned acquisition function operating within a chance-constrained optimization framework outperforms prior work in active learning. Specifically, we are able to simultaneously achieve an improvement in information gain via increased sample efficiency and decreased computation time. We achieve a 46\% increase in information gain while still achieving a 20\% speedup in computation compared to active learning baselines and 60x faster computation time compared to our meta learning baseline. Our novel, deep learning architecture, demonstrates a unique ability to leverage the power of feature learning across time-series data within a LSTM neural network and the utility of deep Q-learning, within mathematical optimization with chance constraints to explicitly tradeoff safety and active learning.
 

 \section{Related Work}
\label{RelatedWorks}
Our work lies at the crossroads of active learning, meta-learning and safe learning. We discuss the contributions of our work and why our approach is novel in light of previous approaches to active learning, meta-learning and safe learning.
 
\textbf{Active Learning - }
 Active learning acquisition functions provide heuristics to selecting the candidate unlabeled training data sample that, if the label were known, would provide the most information to the model being learned \cite{Burbidge2007,Hasenjager1998,Cai2017,Hastie2017}. Snoek et al. \cite{Snoek2012} discusses various Gaussian Process based methods, and Konyushkova et al. \cite{Konyushkova2017} presents a method by which to predict the error reduction from a data sample via training on prior experience. Previous literature has also investigated on-the-fly active learning and meta-active learning \cite{Bachman2016}.  For example, the work by  Geifman and El-Yaniv \cite{Geifman2018} attempts to actively learn the neural network architecture that is most appropriate for the problem at hand and can be integrated with an active learning querying strategy. Kirsch et al. \cite{Kirsch2019} takes a similar approach to Bayesian batch active learning and scores samples based on mutual information. Pang et al. \cite{Pang2018} proposed a method by which to learn an acquisition function which can generalize to a variety of classification problems.  However, this work has only been demonstrated for classification.

\textbf{Meta-Learning for Dynamics - }
Prior work has attempted to address the problem of learning altered dynamics via a meta-learning approach \cite{Clavera2009}. For example Belkhale et al. \cite{Belkhale2020} investigates a meta-learning approach to learn the altered dynamics of an aircraft carrying a payload; the authors train a neural network on prior data to predict various unknown environmental and task factors allowing the model to adapt to new payloads.  Finn et al. \cite{Finn2018} present a meta-learning approach to quickly learning a control policy.  In this approach, a distribution over prior model parameters that are most conducive to learning of the new dynamics is meta-learned offline. While this approach provide fast policies for learning new dynamics, it does not explicitly reason about sample efficiency or safety.  

\textbf{Safe Learning -}
Prior work has investigated safe learning in the context of safe Bayesian optimization and safe reinforcement learning. For example, Yanan et al. \cite{Sui2015} develops the algorithm SafeOpt which balances exploration and exploitation to learn an unknown function. Their algorithm assumes that the unknown function meets a variety of assumptions including that the unknown function meets regularity conditions expressed via a Gaussian process prior and that it is Lipschitz continuous. Turchetta et al. \cite{Turchetta2016} addresses the problem of safely exploring an MDP by defining an a priori unknown safety constraint which is updated during exploration and Zimmer et al. \cite{Zimmer2018} utilizes a Gaussian process for safely learning time series data, including an active learning component to improve sample efficiency. However, neither of these approaches incorporate the knowledge from prior data to increase sample efficiency, limiting their ability to choose the optimal action.

\section{Conclusion} In this paper we demonstrate a state of the art meta-learning approach to active learning for control.  By encoding the context of the dynamics via an LSTM and learning a Q-function -- which we encode into a mixed-integer optimization framework -- that predicts the expected information gain of taking a given action, our framework is able to efficiently and safely learn the nature of the altered dynamics.  We compare our approach to baseline acquisition functions and demonstrate that ours is superior in both computation time and information gain, achieving a 46\% increase in information gain while still achieving a 20\% speedup in computation time compared to state-of-the-art acquisition functions and more than a 58\% higher information compared to Bayesian approaches.

\section{Broader Impact}

\paragraph{Beneficiaries --} This work has far reaching implications in various domains including healthcare and robotics. The ability to learn from small heterogeneous datasets in healthcare and apply this knowledge to future patients has the potential to greatly improve patient care and outcomes. For instance, determining the optimal parameters for a deep brain stimulation device to reduce seizures in epilepsy patients takes extensive time and effort when done manually by a physician. We demonstrate that our algorithm can efficiently and safely determine the optimal setting for deep brain stimulation via learning from past data, thus potentially decreasing the time that a patient must live with seizures. Furthermore, as autonomous robotic systems become more common in our daily lives, there is an increased need to for these systems to be able to safely and efficiently learn the nature of alterations in their dynamics. We demonstrate that our algorithm can safely learn the damaged dynamics of an aircraft in subsecond time while imposing safety constraints to prevent unrecoverable configurations from occurring.
\paragraph{Negatively affected parties --} We do not believe that any party will be negatively impacted by our work. Our work seeks to push the boundaries of safe active learning and to improve upon complementary work in the field.
\paragraph{Implications of failure --} A failure of our method will result in a less than optimal policy when seeking out new data. However, due to our algorithm's ability to impose safety constraints, as long as these are correctly specified by the end user, such a failure would not be catastophic or life-threatening.
\paragraph{Bias and Fairness --} The learned acquisition function will be biased based on the distribution of the offline training data set. This bias may be overcome by diversifying the training data set.




\bibliographystyle{IEEEtran}

\bibliography{bibliography.bib}

\end{document}